\documentclass[journal,twoside,web]{ieeecolor}
\usepackage{generic}
\usepackage{cite}
\usepackage{amsmath,amssymb,amsfonts}
\usepackage{graphicx,multirow}
\usepackage{hyperref}
\usepackage{latexsym}
\usepackage{url}
\usepackage{booktabs}
\usepackage{textcomp} 
\usepackage{lineno}
\usepackage[ruled]{algorithm2e}
\usepackage{algorithmic}
\hypersetup{hidelinks=true}
\usepackage{textcomp}
\def\BibTeX{{\rm B\kern-.05em{\sc i\kern-.025em b}\kern-.08em
    T\kern-.1667em\lower.7ex\hbox{E}\kern-.125emX}}
\begin{document}
\title{Learning to learn skill assessment for fetal ultrasound scanning}
\author{Yipei Wang, Qianye Yang, Lior Drukker, Aris T. Papageorghiou, Yipeng Hu, J. Alison Noble
\thanks{Y.W, Q.Y, Y.H and A.N are with Institute of Biomedical Engineering, Department of Engineering Science, University of Oxford, Oxford, OX3 7DQ, UK (email: yipei.wang@ucl.ac.uk)}
\thanks{Y.W, Q.Y and Y.H are with Department of Medical Physics and Biomedical Engineering, University College London, London, WC1E 6BT, UK.}
\thanks{L.D and A.P are with Nuffield Department of Women's \& Reproductive Health, University of Oxford, Oxford, OX3 9DU, UK.}
\thanks{L.D is with Rabin-Beilinson Medical Center, Tel-Aviv University Faculty of Medicine, Israel}}

\maketitle

\begin{abstract}
Traditionally, ultrasound skill assessment has relied on expert supervision and feedback, a process known for its subjectivity and time-intensive nature. 
Previous works on quantitative and automated skill assessment have predominantly employed supervised learning methods, often limiting the analysis to predetermined or assumed factors considered influential in determining skill levels.
In this work, we propose a novel bi-level optimisation framework that assesses fetal ultrasound skills by how well a task is performed on the acquired fetal ultrasound images, without using manually predefined skill ratings. 
The framework consists of a clinical task predictor and a skill predictor, which are optimised jointly by refining the two networks simultaneously.
We validate the proposed method on real-world clinical ultrasound videos of scanning the fetal head. The results demonstrate the feasibility of predicting ultrasound skills by the proposed framework, which quantifies optimised task performance as a skill indicator.
\end{abstract}


\section{Introduction}
\label{sec:introduction}

Ultrasonography is the most commonly used clinical medical imaging technique for monitoring maternal and fetal well-being throughout pregnancy, assessing fetal growth, and visualising the fetal anatomy, due to its safety, relatively low-cost, real-time and non-invasive nature \cite{salomon2022isuog}. 
During fetal ultrasound scanning, a trained user (sonographer) performs a series of tasks, requiring careful manipulation of the probe while observing the visualised fetal anatomy, locating several specific diagnostic planes based on standard plane definitions and guidelines, and performing biometric measurements ~\cite{drukker2021}. Proficiency in ultrasound is a skill that is hard to achieve and difficult to objectively assess.

Commonly used traditional methods for assessing ultrasound skills are generally manual, subjective and time-consuming, which typically include rating the acquired images using predefined criteria \cite{blehar2015learning, wanyonyi2014image}, and observing and grading trainee performance as they conduct a series of structured tasks \cite{wragg2003assessing}.
Recent efforts have considered automated assessment of ultrasound skills using statistical analysis and machine learning-based methods \cite{sharma2021multi,wang2020differentiating,bell2017sonographic,holden2019machine}. These automated methods have defined skill based on factors such as the number of examinations performed \cite{bell2017sonographic}, ultrasound task performance scores based on a global rating scale \cite{holden2019machine}, and years of practising experience \cite{wang2020differentiating,sharma2021multi}.
These prior works, which are all supervised learning-based methods, rely on predefined skill differentiation ratings as ground-truth training labels, which can be subjective and may not accurately represent the actual skill of a sonographer when performing a specific scan. For example, despite spending the same amount of time on training or practising scanning, operators might possess varying skill levels due to differences in their learning curves.

In this paper, we propose a novel approach using a specific clinical task to assess sonographer skill in fetal ultrasound. This method rates skills by how well a task is performed on the acquired fetal ultrasound images, without using manually predefined skill ratings.
We consider the clinical task performed by a deep learning model, called a \textit{task predictor}, and build a skill assessment model that predicts the performance of the task predictor as an indicator of skill associated with the acquired ultrasound scan. Therefore, training such a \textit{skill predictor} is dependent on the task predictor.

Our previous work \cite{wang2022task} investigated an arguably ``simplified'' scenario in which a fixed, pre-trained task predictor was used. It is considered simplified because using the fixed task predictor feedback to optimise skill predictors, as in ~\cite{wang2022task}, implies the task predictor training does not impact the skill assessment, an independence that can be questioned in many clinical tasks. 
For example, consider a scenario where the downstream task involves segmenting ultrasound frames from which biometrics can be measured~\footnote{In this task, the segmentation can be motivated by several practical considerations such as model explanation and generalisation, as opposed to an end-to-end biometrics prediction. Exploring the benefits of segmentation however is beyond the scope of this work.}. 
In this context, it becomes more important to optimise the task predictor to excel on a subset of these ultrasound frames. This focus ensures that the biometric measurements can be sufficiently made on those frames, which can indicate the level of the associated skill, rather than aiming for superior average segmentation for all frames across the entire scan.
This example illustrates a key insight in the proposed work: determining the subset of frames that are required to be segmented well is dependent on the skill assessment definition.
Consequently, different skill criteria should prompt adjustments in how the task predictor is optimised, which in this example is how the subset should be selected, instead of using a predefined subset-selection criteria which relies on a fixed (therefore independent of the skill predictor training) task predictor as in the previous work.

In this study, we take into account this co-dependency between the task predictor and the skill predictor. The training of the clinical task predictor becomes conditional on the training of the skill predictor.
The skill predictor assigns scores to input data, that act as weighting factors to the loss function for training the task predictor. Simultaneously, the task predictor is trained to perform well on the data which are associated with higher skill scores. 
The optimisation of the skill predictor is dependent on the optimisation of the task predictor, as the performance of the task predictor serves as the supervision of the skill predictor output.
As the optimisation of the skill predictor is conditioned on the task predictor being optimised, we model the solution in a bi-level optimising framework \cite{sinha2017review,liu2021investigating}. 

The contributions of this work are: 1) We propose a bi-level optimisation based framework for assessing ultrasound skills consisting of two jointly-trained neural networks, which takes into account the co-dependency between them to facilitate a class of objective, task-specific skill assessment approaches that have not been investigated before; 2) The method is validated on three different clinically motivated segmentation tasks, using ultrasound images in a real-world fetal ultrasound dataset, acquired from 176 subjects by multiple participating sonographers; and 3) The implementation of this framework is available at: \url{https://github.com/pipiwang/Ultrasound-skill-assessment}.

\section{Method}
\begin{figure}[htb]
     \centering
     \includegraphics[width=0.5\textwidth]{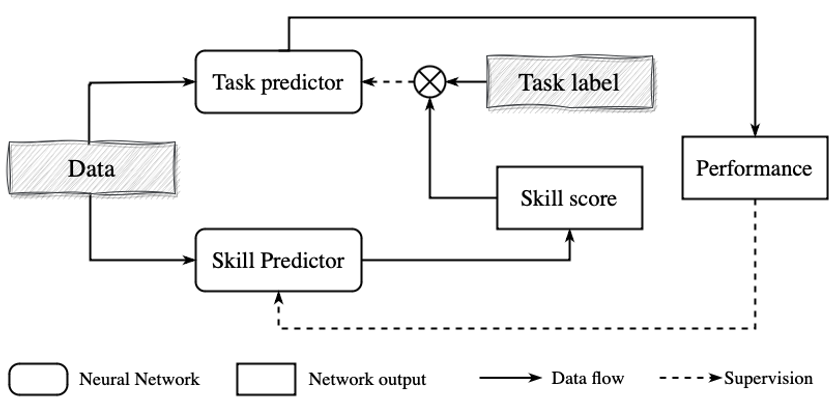}
     \caption{Overview of the proposed method. }
     \label{fig:overview}
 \end{figure}

An overview of the proposed method is shown in Fig.~\ref{fig:overview}.
Assume a total of $I$ subjects, each individually associated with an ultrasound video. Let $V$ denote a sequence of consecutive ultrasound frames randomly sampled from the video of the $i^{th}$ subject, beginning at time point $t$, and $G$ represent the corresponding frame-wise ground-truth label for a specific clinical task within the sequence. Each ultrasound frame sequence is denoted as $V = \left\{ v_j | j=t, t+1,t+2,...t+\tau-1 \right\}$, and its corresponding ground-truth labels are represented as $G = \left\{ g_j | j=t,t+1,t+2,...t+\tau-1 \right\}$, where $v_j$ and $g_j$ denotes the ultrasound image frame and ground-truth label at time index $t$ from the video of the $i^{th}$ subject, and $\tau$ denotes the length of the frame sequence.

\begin{figure*}[htb]
    \centering
    \includegraphics[width=0.9\textwidth]{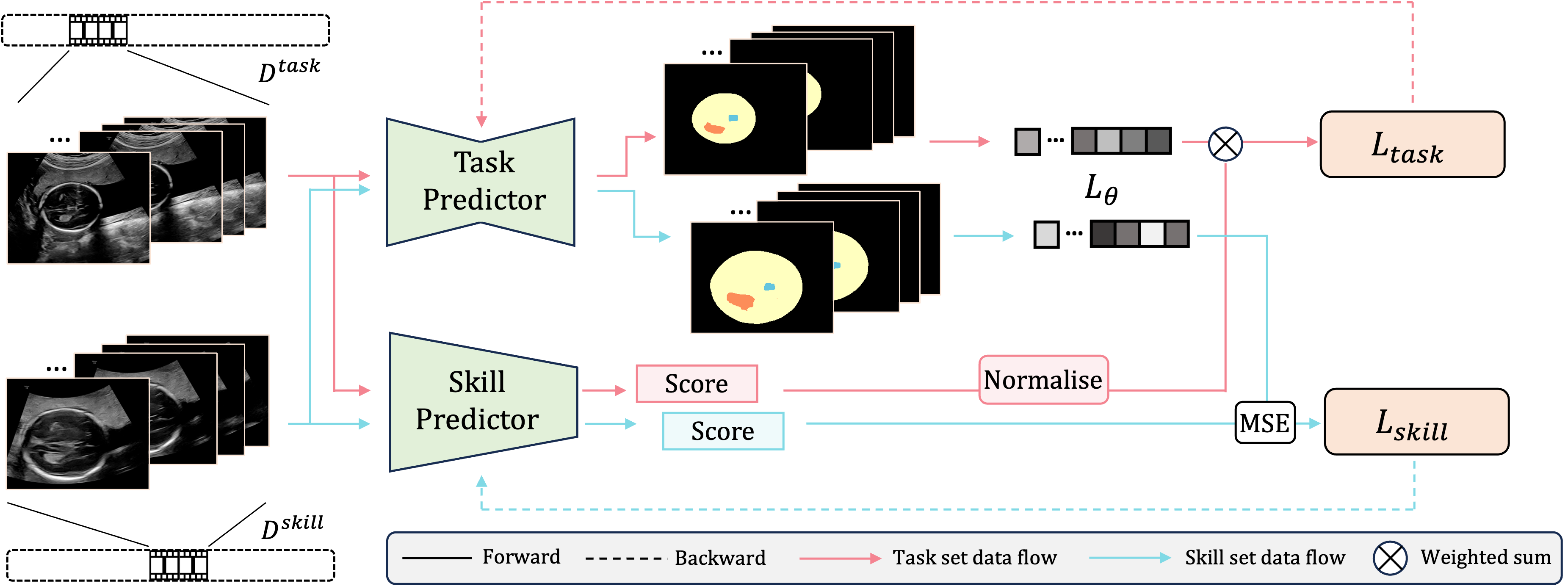}
    \caption{Illustration of the proposed bi-level optimisation framework for ultrasound skill assessment. }
   \label{fig:framework}
\end{figure*}

The dataset $\mathcal{D}^{train}$ used for building the models consists of ultrasound frame sequences and labels from $I$ subjects, given by $\mathcal{D}^{train} = \left\{(V, G) \right\}$. The dataset is randomly partitioned into two sub-datasets, $\mathcal{D}^{train} = \mathcal{D}^{task} \cup \mathcal{D}^{skill}$, where $\mathcal{D}^{task}$ and $\mathcal{D}^{skill}$ denote the training datasets for the task predictor and the skill predictor respectively. $\mathcal{D}^{task} \cap\mathcal{D}^{skill} = \emptyset$ was guaranteed, as widely adopted in bi-level optimisation to reduce over-fitting \cite{sinha2017review}. A detailed illustration of the proposed framework is presented in Fig.~\ref{fig:framework}.

\subsection{Task predictor} 
We consider the clinical task of segmenting the anatomical landmarks on an ultrasound image sequence by assigning each ultrasound frame from the sequence a segmentation mask. 
The clinical task predictor is a deep neural network $f_{task}\left( V;\theta \right): \mathcal{V}\rightarrow \mathcal{G}$ with learnable parameters $\theta$, which takes an ultrasound image sequence $V$ from time point $t$ with the length of $\tau$ as input that is trained to produce segmentation prediction for each frame within the sequence, given by ${\hat G} = f_{task}\left( V;\hat{\theta} \right)$. 
The task predictor is trained using ground-truth labels $G$, which corresponds to each input sequence $V$ of the same dimensions.
It is important to clarify that this is a sequence-based task for our clinical application of interest, but is not required in the proposed skill assessment framework. Other types of tasks such as those based on single-frame as network input should be applicable.

\subsection{Skill predictor}
The skill predictor, denoted as $f_{skill}\left(V;\omega \right): \mathcal{V} \rightarrow \mathcal{S}$, is a second deep neural network with learnable parameters $\omega$. 
The skill predictor takes an ultrasound image sequence $V$ as input, and is trained to predict a skill score ${\hat s} = f_{skill}\left( V;\hat{\omega} \right)$. 
Notably, ${\hat s}$ is a scalar value representing a predicted skill score for an ultrasound image sequence that begins at time point $t$. 
The skill predictor is optimised using $s$, which is derived from an evaluation metric computed on the task predictor. 
The supervision score reflects the performance of the task predictor on the input image sequence $V$, detailed in Sect.\ref{sec:method-opt-skill}.
A higher skill score $\mathcal{S} \in [0,1]$ corresponds to, as defined in this study, better performance in acquiring the input sequence for the downstream task.

\subsection{Optimising the task predictor}
\label{sec:method-opt-task}
Let $L^{\theta}$ denote a loss function that computes the frame-wise difference between the task predictor output $f_{task}\left( V;\theta \right)$ and the target ground-truth label $G$. An example of such a loss function is mean Dice loss, averaged over all the frames in a sequence. The task predictor is optimised by minimising $L^{\theta}$ weighted by the predicted skill score:
\begin{equation}
\label{eq:ltask}
    L_{task} = \mathbb{E}_{(V, G)\sim\mathcal{D}^{task}}\left[L^{\theta}\left( f_{task}\left( V;\theta \right), G\right) \cdot \hat{s}^{norm} \right],
\end{equation}
where $\mathbb{E}( \cdot)$ is the expectation over all training data and $\hat{s}^{normed} $ denotes the predicted skill score normalized within a training minibatch.

The loss function formulation (Eq.~\ref{eq:ltask}) takes into account two important aspects during training. First, by minimising $L^{\theta}$, the network seeks to improve its performance on the target task by aligning the prediction with the target ground-truth label, as it is monotonic to the unweighted $L^{\theta}$. Second, individual network input, here ultrasound sequences, are weighted by the normalised skill score $\hat{s}^{norm}$. The normalisation does not alter the order of the predicted scores. This weighting reflects the relative sequence importance in quantifying skill assessment.

Two different normalisation strategies are considered, within a given training minibatch $\mathcal{D}^{task}_{batch} \subseteq \mathcal{D}^{task}$. 
The min-max-normallised scores $\hat{s}^{norm}_{minmax}$ is given as follows:
\begin{equation}
\label{eq:norm-minmax}
    \hat{s}^{norm}_{minmax}=\frac{\hat{s} -\min_{V\sim\mathcal{D}^{task}_{batch}} \left( \hat{s}\right)}{\max_{V\sim\mathcal{D}^{task}_{batch}} \left( \hat{s}\right) - \min_{V\sim\mathcal{D}^{task}_{batch}} \left( \hat{s}\right)} ,
\end{equation}
where $\min_{V\sim\mathcal{D}^{task}_{batch}} \left( \hat{s}\right)$ and $\max_{V\sim\mathcal{D}^{task}_{batch}} \left( \hat{s}\right)$ represent the minimum and maximum skill scores within the minibatch, respectively. The min-max normalisation linearly scales the predicted skill scores in a minibatch, preserving the original score distribution.
Let $r$ denote the rank of the given skill score within the minibatch, the rank-based normalisation thus is: 
\begin{equation}
\label{rank}
    \hat{s}^{norm}_{rank} = \frac{r-1}{N-1},
\end{equation}
where $1\leq r \leq N$ and a larger value of $r$ indicates a higher skill score. The rank normalisation does not use the exact values of the predicted skill scores, emphasising their relative order in a minibatch.

The normalisation could be parameterised with learnable weights, such as a linear scaling and a shifting \cite{ioffe2015batch}, estimated from the entire train set $\mathcal{D}^{task}$. However it was found practically sufficient to estimate both linear $\hat{s}^{norm}_{minmax}$ and nonlinear $\hat{s}^{norm}_{rank}$ within a minibatch $\mathcal{D}^{task}_{batch}$ (as reported in Sect.~\ref{sec:results}). 

\subsection{Optimising the skill predictor}
\label{sec:method-opt-skill}
Let $L_{mse}$ denote the mean-squared-error loss. The skill predictor is optimised using the following loss function:
\begin{equation}
\label{eq:lskill}
    \begin{split}
    L_{skill} =\mathbb{E}_{(V, G)\sim\mathcal{D}^{skill}} [ L_{mse} ( f_{skill}( V;\omega ), L^{\theta}( f_{task}( V;\theta ), G) )],   
    \end{split}
\end{equation}
By minimising the skill loss $L_{skill}$, the output of the skill predictor which is a skill score for an input sequence, approximates the true performance of the task predictor on chosen samples from the image sequence.

The formulation of $L^{\theta}$ varies depending on the clinical task and the definition of skill. In this work, we propose three definitions, which represent different aspects of skill performance. 
\begin{itemize}
    \item $L^{\theta}_{min} = \min_{j \sim \{t,...,t+\tau-1\}} \{l_{dice}(\hat{g}_j, g_j))\}$, the best performance.
    \item $L^{\theta}_{avg} = \frac{1}{\tau}\sum_{j=t}^{t+\tau-1} \{l_{dice}(\hat{g}_j, g_j)) \}$, the average performance.
    \item $L^{\theta}_{top-m\%} = \frac{100}{m \cdot \tau}\sum_{j \in M} \{l_{dice}(\hat{g}_j, g_j)\} $, the average performance of the top $m\%$ image frames, where $M$ denotes the sequence indices of these frames. 
\end{itemize}
where $l_{dice}$ is the frame-wise Dice loss and, during training, $V$ and $G$ are randomly sampled from each of the subjects in the skill-predictor-training dataset $(V, G)\in \mathcal{D}^{skill}$.

\subsection{Bi-level optimisation of the networks}
We jointly optimise the parameters of the task predictor $\theta$ and the skill predictor $\omega$. Thus the learning problem can be formulated as follows,
\begin{equation}
    \begin{split}
    & \omega^{\ast} = \arg \min_{\omega} \mathbb{E}_{(V, G)\sim\mathcal{D}^{skill}} [ L_{mse} ( f_{skill}( V;\omega ),  L^{\theta}( f_{task}( V;\theta^{\ast} ), G) )],\\
    & \text{s.t. } \theta^{\ast} =  \arg \min_{\theta} \mathbb{E}_{(V, G)\sim\mathcal{D}^{task}}\left[L^{\theta}\left( f_{task}\left( V;\theta \right), G\right) \cdot \hat{s}^{norm} \right]
    \end{split}
\end{equation}
With Eqs.~\ref{eq:ltask} and~\ref{eq:lskill}, the above optimisation can be simplified, with respect to the learnable parameters at the two levels, as follows, 
\begin{equation}
    \begin{split}
        & \omega^{\ast} = \arg \min_{\omega} L_{skill}(\omega, \theta^{\ast}(\omega)), \\
        & \text{s.t. } \theta^{\ast}(\omega) = \arg \min_{\theta} L_{task}(\omega, \theta).
    \end{split}
\end{equation}
Both loss functions $L_{skill}$ and $L_{task}$ are determined not only by parameters $\omega$ for the skill predictor $f_{skill}$, but also by parameters $\theta$ for the task predictor $f_{task}$.
During the optimisation process, the skill predictor aims to find optimal network parameters $\omega$ that minimise loss $L_{skill}(\omega^{\ast}, \theta^{\ast})$, while conditioned on that the weights $\theta$ from the task predictor minimising the loss $L_{task}(\omega^{\ast}, \theta)$.
This conditioned optimisation process is a bi-level optimisation problem \cite{sinha2017review,liu2021investigating}. The skill predictor loss $f_{skill}$ and the task predictor loss $f_{task}$ serve as the upper-level and lower-level objectives, while the parameters $\omega$ and $\theta$ are the upper-level and lower-level parameters, respectively. 
In this work, we use a gradient-based method for this bi-level optimisation problem. 
For each training step, instead of optimising the lower-level objective $ \theta^{\ast}(\omega) = \arg \min_{\theta} L_{task}(\omega, \theta) $ until converge, only one gradient-update step is used to approximate $\theta^{\ast}$, which is a common strategy used in bi-level optimisation~\cite{liu2018darts}, formulated as:
\begin{equation}
\label{eq:grad}
    \nabla_{\omega}L_{skill}(\omega, \theta^{\ast}(\omega)) \approx \nabla_{\omega}L_{skill}(\omega, \theta-\alpha L_{task}(\omega, \theta)),
\end{equation}
where $\alpha$ denotes the learning rate of the lower-level training process.
Applying the chain rule to the approximated gradient in Eq.~\ref{eq:grad} will introduce a computationally expensive and potentially unstable Hessian term. To avoid the higher-order derivative estimation, we assume that for each upper-level training step, the current lower-level model is optimal, such that $\theta = \theta^{\ast}(\omega)$, thus the learning rate $\alpha = 0$, therefore: 
\begin{equation}
    \nabla_{\omega}L_{skill}(\omega, \theta^{\ast}(\omega)) \approx \nabla_{\omega}L_{skill}(\omega, \theta).
\end{equation}
In such case, the second-order derivative is replaced by a first-order approximation, which empirically results in significant computational speedup while maintaining comparable performance, as reported in \cite{liu2018darts, he2020milenas}.
As a result, the bi-level optimisation problem is solved by alternately optimising the upper-level $f_{skill}$ and the lower-level $f_{task}$, with a detailed algorithm described in Alg.~\ref{alg}.

\begin{algorithm}
\caption{}\label{alg}
\SetKwInOut{Input}{Input}
\SetKwInOut{Output}{Output}
\Input{$\mathcal{D}^{task},\mathcal{D}^{skill}$}
\Output{Skill predictor $f_{skill}(\cdot;\omega)$}
\While {not converged}{
\For{Each training epoch}{
    \For{Each training step $k \gets 1$ to $K$}{
        Sample a minibatch $({V}_{skill}^k,{G}_{skill}^k)\sim \mathcal{D}^{skill}$;\\
        
        Compute the output of skill predictor ${\hat{s}}_{skill}^k = f_{skill}({V}_{skill}^k;\omega^k)$;\\
        
        Compute the performance of $f_{task}(\cdot;\theta^k)$ on this minibatch ${s}_{skill}^k = L^{\theta}\left( f_{task}\left( {V}_{skill}^k;\theta^{k} \right), {G}_{skill}^k\right)$;\\
        
        Calculate $L_{skill}^k({\hat{s}}_{skill}^k,{s}_{skill}^k)$;\\
        
        Update the skill predictor weights $\omega^k \gets \omega^{k+1}$ using $\nabla_{\omega} L_{skill}^k(\omega,\theta^k)$\\
        
        Sample a minibatch from $({V}_{task}^k,{G}_{task}^k)\sim \mathcal{D}^{task}$;\\
        
        Compute the output of task predictor ${\hat{G}}_{task}^k = f_{task}({V}_{skill}^k;\theta^k)$;\\
        
        Compute the predicted performance score ${\hat{s}}_{task}^k = f_{skill}({V}_{task}^k;\omega^{k+1})$ and normalise within the minibatch $\hat{s}^{norm}$;\\
        
        Calculate $L_{task}^k=L^{\theta}({\hat{G}}_{task}^k,{G}_{task}^k)\cdot\hat{s}^{norm}$; \\
        
        Update the task predictor weights $\theta^k \gets \theta^{k+1}$ using $\nabla_{\theta} L_{task}^k(\omega^{k+1},\theta)$\\
    }
}
}
\end{algorithm}

\section{Experiments}
\subsection{Model selection}
The task predictors and skill predictors were jointly trained as a bi-level optimisation model on the training dataset $\mathcal{D}^{train}$ for a predefined number of epochs. Recall that $\mathcal{D}^{train} = \mathcal{D}^{task} \cup \mathcal{D}^{skill}$, where $\mathcal{D}^{task}$ and $\mathcal{D}^{skill}$ were used for training the task predictor and the skill predictor respectively. 
After training, we selected the best-performing model by evaluating the performance on the skill predictor network. Specifically, we measured the mean squared error (MSE) of the skill predictor on dataset $\mathcal{D}^{task}$ and the model with the lowest MSE was selected for evaluation. Furthermore, through our empirical observations, the specific task predictor considered in this paper typically reached convergence after $500$ training epochs. As a result, the evaluation model was selected after $500$ epochs of training.

\subsection{Direct evaluation}
For direct evaluation, the skill predictor and task predictor from the selected evaluation model were evaluated on the testing dataset separately.
Note that the skill scores generated by the task predictor were also utilised as supervision scores for assessing the performance of the skill predictor.
Direct evaluation provides insights into model generalisation ability to unseen data during training without adaptation.

\subsection{Meta learning evaluation}
The second evaluation method involved a meta-learning evaluation procedure, which was used to imitate real-world scenarios where the model need to be adapted before applying to new data. For brevity, we term this evaluation process as meta evaluation in the following sections. Meta evaluations were performed following three steps.
\subsubsection{Meta evaluation dataset}
The test dataset was partitioned into two sub-datasets, a meta-evaluation train set which was used for fine-tuning the selected model and a meta-evaluation test set for the final testing. In this work, we split the test dataset using incremental proportions of the data for training, ranging from $10\%$ to $60\%$.
\subsubsection{Fine-Tuning on meta-evaluation train dataset} 
The selected model was fine-tuned using the meta-evaluation train dataset. We selected the fine-tuned models at intervals of $10$ epochs, ranging from $10$ to $100$ epochs for evaluation.
\subsubsection{Testing on meta-evaluation test dataset}
The selected fine-tuned models were evaluated on the meta-evaluation test set. Similar to the direct evaluation, both the task predictor and the skill predictor were tested separately. 

\subsection{Clinical task and dataset curation}
Ultrasound videos used in this work came from the PULSE study~\cite{drukker2021}. The PULSE study was approved by the UK Research Ethics Committee under Reference 18/WS/0051. Second trimester ultrasound scans were performed by qualified sonographers using a commercial Voluson E8 version BT18 (General Electric Healthcare, Zipf, Austria) ultrasound machine with a standard curvilinear (C2-9-D, C1-5-D) and 3D/4D (RAB6-D) probes. Written informed consent was given by all participating pregnant women.
In this work, we are interested in the anatomical landmarks that identify the head circumference measurement plane (HCP). Head circumference measurement is a clinical task performed during routine second-trimester fetal ultrasound scanning for monitoring fetal growth and gestational age estimation. For the experiments in our study, a subset of the PULSE dataset was selected with the following criteria:
\begin{enumerate}
    \item The sonographer is searching for an optimal HCP view and the scanner is not in image freeze mode during the video clip;
    \item the video clip is before the freeze frame that has been assigned an anatomy view of ``HC'' (the head circumference measurement) identified by optical character recognition;
    \item the video clip has a length of at least 6 seconds, approximately $342 \pm 42$ frames.
\end{enumerate}
We extracted ultrasound videos from up to $12$ seconds before the time that the sonographer stops scanning to measure the head circumference of each ultrasound scan.

\begin{figure}
     \centering
     \includegraphics[width=0.5\textwidth]{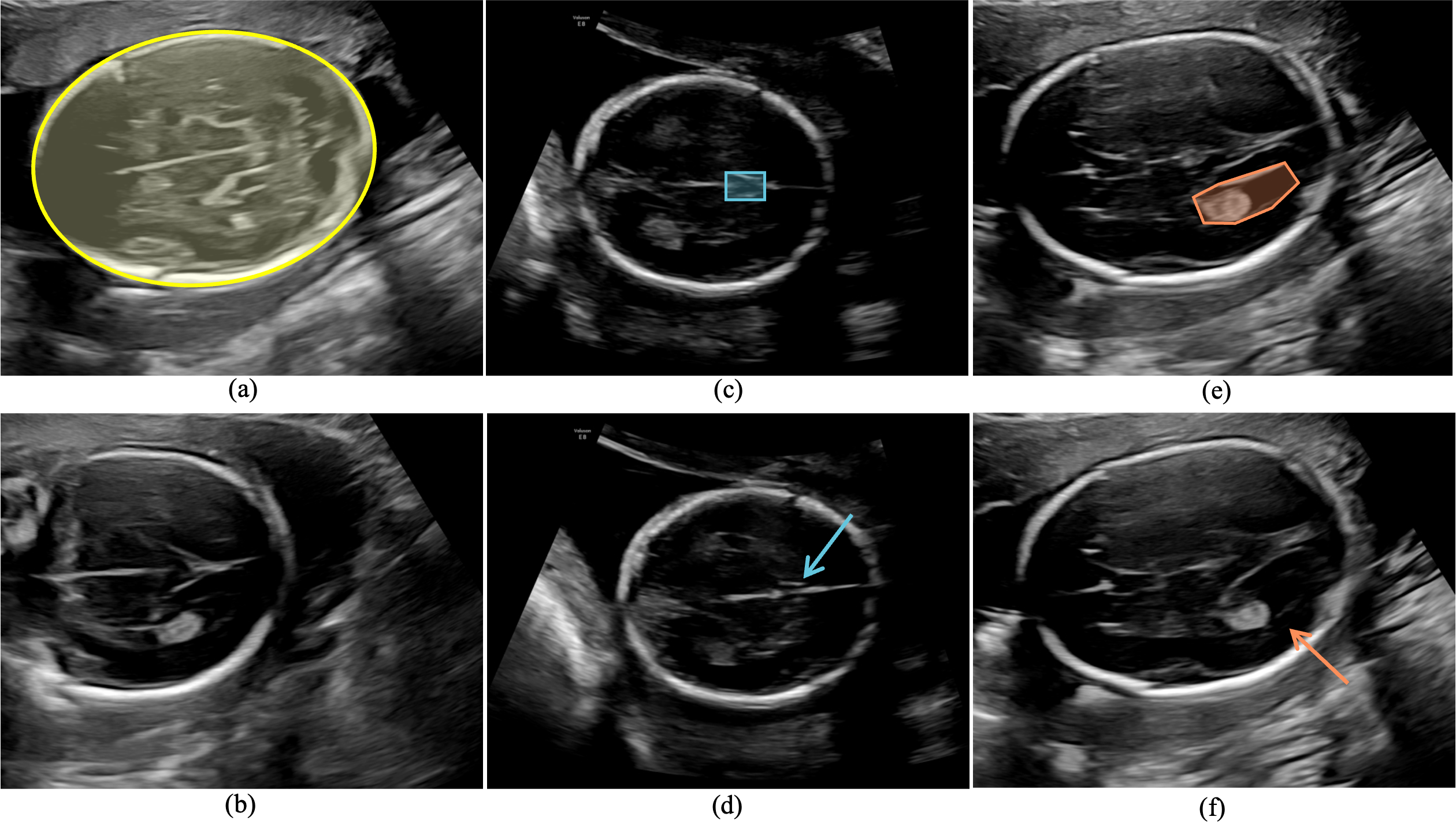}
     \caption{Example frames with and without annotations for the anatomical landmarks. }
     \label{fig:label}
 \end{figure}

\begin{table}[]
\centering
\caption{Dataset description}
\label{tab:dataset}
\resizebox{0.25\textwidth}{!}{%
\begin{tabular}{lllll}
\toprule
\multirow{2}{*}{} & \multicolumn{2}{l}{$\mathcal{D}^{train}$} & \multicolumn{2}{l}{$\mathcal{D}^{test}$} \\ \cline{2-5} 
                  & Frames        & Scans        & Frames        & Scans       \\ \midrule
HC                & 44114         & -            & 6345          & -           \\
CSP               & 14242         & -            & 1964          & -           \\
LV                & 10453         & -            & 716           & -           \\
Total             & 60299         & 151          & 8353          & 25         \\ \bottomrule
\end{tabular}%
}
\end{table}

During a second trimester scan, a transventricular plane (TVP) is obtained to assess the anatomical integrity of the fetal head \cite{gallery2020isuog} and for fetal head biometry \cite{napolitano2016scientific}.
For a standard tranventricular plane, there should be a clear midline echo broken by the cavum septi pellucidi (CSP) and the presence of the lateral ventricle (LV) with its atrium. 
We selected and manually annotated three anatomical landmarks that are visible in the transventricular plane following the ISUOG guidelines \cite{salomon2011practice,gallery2020isuog} and NHS FASP handbook \cite{fasp}.
All frames from the video were densely annotated manually using a local Linux version of the Computer Vision Annotation Tool (CVAT) \cite{boris_sekachev_2019_3497106}. 
Prior to the annotation process, the annotator received 6 years of biomedical imaging training, including 3 years of experience in identifying and interpreting fetal ultrasound anatomical structures. 
The detailed annotation process for each ground-truth segmentation structure mask is described next.

\textbf{Head circumference} (HC) was annotated by fitting an ellipse on the outer skull boundary. 
Note that for HC annotation, we did not use the traditional definition of the HC where it is only obtained by the standard plane, an ellipse was drawn whenever an intact cranium is shown on the frame. Figure~\ref{fig:label}-(a) and (b) present example frames from the same clip where (b) shows a partial skull due to the obstructed view caused by angle and position of the probe. An ellipse was annotated on Fig.~\ref{fig:label}-(a) because of the appearance of an intact skull although it is not a standard tranventricular view.

\textbf{Cavum septi pellucidi} (CSP) is a cavity filled with fluid with two thin membranes which appear as two white lines with one dark box in between. The CSP was annotated by fitting a rectangle on the frame where the CSP appears. Figure~\ref{fig:label}-(c) shows an example CSP annotated by a rectangle. Note that the midline echo is absent in the centre of the CSP, whereas in some of the similar views (as shown in Fig.~\ref{fig:label}-(d)), the midline echo appears which indicates a possible finding of fornix instead of CSP \cite{winter2010cavum}.

\textbf{Lateral ventricle and the atrium} (LV) is usually complex in shape and we annotated it with a free polygon. As illustrated in Fig.~\ref{fig:label}-(e), the atrium can be identified by the highly echogenic choroid plexus which is surrounded by the non-echogenic fluid that shows the boundaries of the atrium of the ventricle. As shown in Fig.~\ref{fig:label}-(e) and (f), the LV was annotated with a polygon when both sides of the ventricle are visible.

About 5\% of the annotations which include 10 ultrasound scans were checked by two qualified sonographers and adjustments to the annotation were made following expert feedback.

The dataset was partitioned into the train set and test set. The train set is randomly split into two subsets, $\mathcal{D}^{task}$ and $\mathcal{D}^{skill}$. Video clips from the same sonographer or the same subject were held in the same subset to prevent data leakage.
The total number of scans and a detailed number of each anatomical landmark annotated are presented in Tab.~\ref{tab:dataset}.
 
\subsection{Model implementation}
All ultrasound frames were resampled to $224 \times 288$ pixels and intensity normalised to a mean of zero and standard deviation of one. A U-Net~\cite{ronneberger2015u} was employed as the anatomical landmark segmentation task predictor. All ultrasound frames from a sequence were consecutively stacked to form the U-Net model input. The generalised soft Dice loss~\cite{sudre2017generalised} was used as $L^{\theta}$ for training the task predictor. For the skill predictor, we modified the ResNet18 \cite{he2016deep} network architecture by adding a sigmoid function layer before the final output. Both networks were implemented in PyTorch version 2.0.1. The models were trained on NVIDIA Quadro GV100 GPUs and used an Adam optimiser with a learning rate of $10^{-4}$.

\subsection{Evaluation metrics}
\subsubsection{Task predictor}
The average Dice score is calculated across all testing ultrasound frames to assess the performance of the segmentation task predictor. 
It is important to note that for the anatomical landmark segmentation task chosen in this work, ground-truth annotations may not exist for all frames. This is because most of the ultrasound frames in the video are in the process of searching for the desired view and only a few of them are planes suitable for diagnosis. In the cases where the ground-truth landmarks are not present, calculating the Dice score can result in a value of $0$ which does not provide a proper assessment of the segmentation model performance. Therefore, we report the Dice score when ground-truth annotations for individual landmarks are available. 

In addition, we report the average symmetric surface distance (ASSD)~\cite{heimann2009comparison} between the predicted masks and the ground-truth annotations. As we do not have information on pixel spacing, we report ASSD in pixel units.

\begin{table*}[htb]
\centering
\caption{Task predictor performance with different normalisation methods and formulations of $L^{\theta}$. Best values are marked in bold.}
\label{res-seg-main}
\resizebox{0.75\textwidth}{!}{%
\begin{tabular}{llllllll}
\toprule
\multicolumn{1}{l}{\multirow{2}{*}{Norm}} & \multicolumn{1}{l}{\multirow{2}{*}{$L^{\theta}$}} &                          \multicolumn{3}{l}{Dice}                      & \multicolumn{3}{l}{ASSD}                           \\ \cline{3-8} 
\multicolumn{1}{l}{}                           & \multicolumn{1}{l}{}                           &  HC             & CSP            & LV             & HC             & CSP            & LV               \\ \midrule
rank                                           & $L^{\theta}_{avg}$                                            &  \textbf{0.971 $\pm$ 0.033} & \textbf{0.813 $\pm$ 0.129} & \textbf{0.912 $\pm$ 0.034} & \textbf{1.828 $\pm$ 2.184} & \textbf{1.691 $\pm$ 1.509} & \textbf{1.942 $\pm$ 2.163}   \\
rank                                           & $L^{\theta}_{min}$                                            &  0.930 $\pm$ 0.137 & 0.669 $\pm$ 0.278 & 0.760 $\pm$ 0.121 & 5.223 $\pm$ 9.919 & 3.296 $\pm$ 5.681 & 10.044 $\pm$ 7.148  \\
rank                                           &$L^{\theta}_{top-20\%}$                                       &  0.969 $\pm$ 0.034 & 0.800 $\pm$ 0.201 & 0.909 $\pm$ 0.037 & 2.062 $\pm$ 2.563 & 1.996 $\pm$ 3.276 & 3.141 $\pm$ 5.429   \\ 
min-max                                            & $L^{\theta}_{avg}$                                            &  0.969 $\pm$ 0.034 & 0.775 $\pm$ 0.207 & 0.827 $\pm$ 0.079 & 2.126 $\pm$ 2.812 & 1.735 $\pm$ 1.880 & 6.501 $\pm$ 3.077   \\
min-max                                            & $L^{\theta}_{min}$                                            &  0.963 $\pm$ 0.038 & 0.691 $\pm$ 0.285 & 0.803 $\pm$ 0.084 & 2.477 $\pm$ 2.556 & 3.383 $\pm$ 4.640 & 7.209 $\pm$ 4.061   \\
min-max                                            & $L^{\theta}_{top-20\%}$                                       &  0.953 $\pm$ 0.064 & 0.678 $\pm$ 0.282 & 0.752 $\pm$ 0.216 & 3.206 $\pm$ 3.832 & 3.398 $\pm$ 5.762 & 10.660 $\pm$ 13.355 \\ \bottomrule
\end{tabular}%
}
\end{table*}

\begin{table*}[htb]
\centering
\caption{Comparitative results for task predictor with different minibatch sizes. Best values are marked in bold. $\ast$ indicates statistically significant compared to the second-place method ($p<0.05$)}
\label{res-seg-bs}
\resizebox{0.8\textwidth}{!}{%
\begin{tabular}{llllllll}
\toprule
\multicolumn{1}{l}{\multirow{2}{*}{Norm}} & \multicolumn{1}{l}{\multirow{2}{*}{\begin{tabular}[c]{@{}c@{}}minibatch \\ size\end{tabular}}}          & \multicolumn{3}{l}{Dice}                      & \multicolumn{3}{l}{ASSD}                               \\ \cline{3-8} 
\multicolumn{1}{l}{}                           & \multicolumn{1}{l}{}                 & HC             & CSP            & LV             & HC               & CSP              & LV               \\ \midrule
rank                                           & 8                                               & \textbf{0.974 $\pm$ 0.028}$\ast$ & \textbf{0.859 $\pm$ 0.082}$\ast$ & \textbf{0.928 $\pm$ 0.034}$\ast$ & \textbf{1.561 $\pm$ 1.754}$\ast$   & \textbf{1.198 $\pm$ 1.027}$\ast$   & \textbf{1.168 $\pm$ 1.628}$\ast$   \\
rank                                           & 16                                              & 0.930 $\pm$ 0.137 & 0.669 $\pm$ 0.278 & 0.760 $\pm$ 0.121 & 5.223 $\pm$ 9.919   & 3.296 $\pm$ 5.681   & 10.044 $\pm$ 7.148  \\
rank                                           & 32                                              & 0.837 $\pm$ 0.235 & 0.585 $\pm$ 0.286 & 0.712 $\pm$ 0.208 & 15.927 $\pm$ 24.291 & 9.673 $\pm$ 12.551  & 11.885 $\pm$ 9.884  \\
rank                                           & 64                                              & 0.894 $\pm$ 0.164 & 0.141 $\pm$ 0.134 & 0.425 $\pm$ 0.287 & 10.677 $\pm$ 16.825 & 28.737 $\pm$ 13.084 & 23.722 $\pm$ 11.647 \\ \midrule
min-max                                        & 8                                               & 0.946 $\pm$ 0.080 & 0.690 $\pm$ 0.294 & \textbf{0.836} $\pm$ 0.100$\ast$ & 3.334 $\pm$ 4.564   & \textbf{1.798 $\pm$ 1.749}$\ast$   & \textbf{4.576 $\pm$ 3.301}$\ast$   \\
min-max                                        & 16                                              & 0.963 $\pm$ 0.038 & 0.691 $\pm$ 0.285 & 0.803 $\pm$ 0.084 & 2.477 $\pm$ 2.556   & 3.383 $\pm$ 4.640   & 7.209 $\pm$ 4.061   \\
min-max                                        & 32                                              & \textbf{0.964} $\pm$ 0.044$\ast$ & \textbf{0.738} $\pm$ 0.247$\ast$ & 0.806 $\pm$ 0.158 & 2.393 $\pm$ 3.133$\ast$   & 2.096 $\pm$ 2.699   & 6.753 $\pm$ 9.773   \\
min-max                                        & 64                                              & 0.672 $\pm$ 0.283 & 0.094 $\pm$ 0.098 & 0.234 $\pm$ 0.230 & 29.893 $\pm$ 27.591 & 31.270 $\pm$ 13.115 & 34.177 $\pm$ 16.476 \\
\bottomrule
\end{tabular}%
}

\end{table*}

\begin{table*}[!h]
\centering
\caption{Task predictor evaluation for different $L^{\theta}_{top-m\%}$. Best values are marked in bold. $\ast$ indicates significance compared to the second-place method ($p<0.05$).}
\label{res-seg-top}
\resizebox{0.8\textwidth}{!}{%
\begin{tabular}{llllllll}
\toprule
\multicolumn{1}{c}{\multirow{2}{*}{Norm}} & \multicolumn{1}{c}{\multirow{2}{*}{$L^{\theta}_{top-m\%}$}} & \multicolumn{3}{l}{Dice}    & \multicolumn{3}{l}{ASSD}                               \\ \cline{3-8} 
\multicolumn{1}{c}{}                           & \multicolumn{1}{c}{}                                      & HC             & CSP            & LV             & HC               & CSP              & LV               \\ \midrule
rank  & $L^{\theta}_{top-20\%}$   &  0.969 $\pm$ 0.034 &  0.800 $\pm$ 0.201 &  0.909 $\pm$ 0.037 &  2.062 $\pm$ 2.563 &  1.996 $\pm$ 3.276 &  3.141 $\pm$ 5.429  \\ 
        rank  & $L^{\theta}_{top-30\%}$  &  \textbf{0.971 $\pm$ 0.030}$\ast$ &  \textbf{0.818 $\pm$ 0.146}$\ast$ &  0.910 $\pm$ 0.044 &  \textbf{1.836 $\pm$ 2.096}$\ast$ &  1.779 $\pm$ 1.962 &  \textbf{1.366 $\pm$ 1.337}  \\
        rank  & $L^{\theta}_{top-40\%}$  &  0.969 $\pm$ 0.034 &  0.785 $\pm$ 0.166 &  0.821 $\pm$ 0.082 &  2.122 $\pm$ 2.817 &  \textbf{1.655 $\pm$ 1.400}$\ast$ &  6.024 $\pm$ 2.866  \\ 
        rank & $L^{\theta}_{top-50\%}$ & 0.967 $\pm$ 0.036 &  0.807 $\pm$ 0.143 &  \textbf{0.916 $\pm$ 0.035}$\ast$ &  2.232 $\pm$ 2.778 &  2.260 $\pm$ 2.491 &  1.507 $\pm$ 1.447 \\ \midrule
        min-max & $L^{\theta}_{top-20\%}$  & 0.953 $\pm$ 0.064 & 0.678 $\pm$ 0.282 & 0.752 $\pm$ 0.216 & 3.206 $\pm$ 3.832 & 3.398 $\pm$ 5.762 & 10.660 $\pm$ 13.355  \\ 
        min-max & $L^{\theta}_{top-30\%}$  & \textbf{0.969 $\pm$ 0.031}$\ast$ & \textbf{0.806 $\pm$ 0.145} & \textbf{0.912 $\pm$ 0.040}$\ast$ & \textbf{1.888 $\pm$ 2.051}$\ast$ & \textbf{1.743} $\pm$ 2.445 & \textbf{2.001 $\pm$ 2.383}$\ast$  \\ 
        min-max & $L^{\theta}_{top-40\%}$  & 0.966 $\pm$ 0.037 & 0.723 $\pm$ 0.263 & 0.850 $\pm$ 0.073 & 2.536 $\pm$ 2.970 & 1.894 $\pm$ 2.403 & 5.227 $\pm$ 3.863  \\ 
        min-max & $L^{\theta}_{top-50\%}$ & 0.965 $\pm$ 0.044 & 0.800 $\pm$ 0.189 & 0.895 $\pm$ 0.084 & 2.230 $\pm$ 3.296 & 1.787 $\pm$ 2.320 & 4.699 $\pm$ 12.221 \\
\bottomrule
\end{tabular}%
}
\end{table*}

\begin{table*}[!h]
\centering
\caption{Effect of sequence length and sampling rate on the task predictor performance. Best values are marked in bold. }
\label{res-seg-seqlen}
\resizebox{0.75\textwidth}{!}{%
\begin{tabular}{llllllll}
\toprule
\multicolumn{1}{c}{\multirow{2}{*}{\begin{tabular}[c]{@{}c@{}}Frame rate \\ (fps)\end{tabular}}} & \multicolumn{1}{c}{\multirow{2}{*}{$\tau$}} & \multicolumn{3}{l}{Dice}                      & \multicolumn{3}{l}{ASSD}                               \\ \cline{3-8} 
\multicolumn{1}{c}{}                                                                             & \multicolumn{1}{c}{}            & HC             & CSP            & LV             & HC               & CSP              & LV               \\ \midrule
30                                                                                               & 10                                           &  0.963 $\pm$ 0.038 & 0.691 $\pm$ 0.285 & 0.803 $\pm$ 0.084 & 2.477 $\pm$ 2.556   & 3.383 $\pm$ 4.640   & 7.209 $\pm$ 4.061 \\
30                                                                                               & 20                                           &  0.921 $\pm$ 0.111 & 0.648 $\pm$ 0.305 & 0.714 $\pm$ 0.197 & 6.267 $\pm$ 8.331   & 2.959 $\pm$ 4.640   & 10.645 $\pm$ 9.668  \\
30                                                                                               & 30                                           & 0.960 $\pm$ 0.056 & \textbf{0.771 $\pm$ 0.164} & 0.882 $\pm$ 0.057 & 2.308 $\pm$ 2.767   & \textbf{1.808} $\pm$ 1.511   & 1.535 $\pm$ 0.888   \\
15                                                                                               & 10                                           &  \textbf{0.966} $\pm$ 0.042 & \textbf{0.771} $\pm$ 0.226 & \textbf{0.892} $\pm$ 0.087 & \textbf{2.178} $\pm$ 2.666   & 1.839 $\pm$ 2.757   & \textbf{1.382 $\pm$ 0.805}   \\
15                                                                                               & 20                                           &  0.953 $\pm$ 0.063 & 0.679 $\pm$ 0.278 & 0.704 $\pm$ 0.177 & 2.748 $\pm$ 3.760   & 1.956 $\pm$ 1.327   & 8.141 $\pm$ 5.033   \\
10                                                                                               & 10                                           &  0.953 $\pm$ 0.066 & 0.653 $\pm$ 0.266 & 0.771 $\pm$ 0.164 & 3.288 $\pm$ 4.964   & 2.399 $\pm$ 2.024   & 6.757 $\pm$ 4.798   \\
10                                                                                               & 20                                           &  0.944 $\pm$ 0.071 & 0.610 $\pm$ 0.291 & 0.714 $\pm$ 0.156 & 3.347 $\pm$ 4.060   & 4.127 $\pm$ 6.134   & 8.105 $\pm$ 2.947  \\ \bottomrule
\end{tabular}%
}
\end{table*}

\subsubsection{Skill predictor}
The MSE between the skill score predicted by the skill predictor and the skill score produced from the segmentation task predictor is reported to evaluate the performance of the skill predictor network.

For a more intuitive evaluation, we are also interested in that, within a given ultrasound scanning, whether the frames being assigned with high skill scores by the skill predictor
would result in improved performance when these higher-rated sequences are input into the task predictor, compared to the average performance obtained when considering the entire scan.
In other words, we investigate whether leveraging the frame sequences most conducive to the task, as identified by the skill predictor, could enhance the practical predictive capabilities of the task predictor.
Given an ultrasound scan, if the task predictor performance on higher-rated frame sequences is better than the average on the entire scan, we denote this scan as a scan with performance improvement.
We calculate the ratio of the number of scans with performance improvement achieved when using the top $1$ or top $5$ rated frame sequences to the number of the total tested scans, denoted as $R$. 
More formally, denote the number of scans with performance improvement and the total number of scans for testing as $N_{improved}$ and $N_{test}$, respectively, and the metric can be formulated as
\begin{equation}
    R = \frac{N_{improved}}{N_{test}} \times 100\%.
\end{equation}
We refer to $R_{top1}$ and $R_{top5}$ as the ratio $R$ when using top 1 and top 5 rated frame sequences respectively.
For top 1, the performance improvement is compared between the task predictor performance on the highest-rated sequence and the average performance on all sequences from the scan; and for top 5, we compare the average task predictor performance on the 5 highest-rated sequences and the average performance on all sequences from the scan.

\subsection{Ablation and comparison studies}
To quantify the impact of different training strategies, designs of objective functions, and data sampling methods on the skill assessment results, the following experiments were performed.

\begin{enumerate}
    \item Two score normalisation methods within a minibatch introduced in Sect.~\ref{sec:method-opt-task} were compared, rank normalisation and min-max normalisation.
    \item Three formulations of $L^{\theta}$ used for optimising the skill predictor, as described in Sect.~\ref{sec:method-opt-skill}, were compared. Specifically, for $L^{\theta}_{top-m\%}$, we performed comparison studies on $m\%$ from $\{20\%, 30\%, 40\%, 50\%\}$.
    \item Comparisons on minibatch size were performed on models using both rank and min-max for normalisation and using $L^{\theta}_{min}$, with a minibatch size of $\{8,16,32,64\}$.
    \item The impact of different sampling rates and lengths of the input frames were also compared.
\end{enumerate}

We also compare our proposed method to current commonly adopted supervised learning scheme. Previous automatic methods are generally supervised approaches that use different criteria for assessing ultrasound skills \cite{holden2019machine, wang2020differentiating, tyrrell2021ultrasound}, therefore, existing methods may not be directly comparable to our proposed method, which do not rely on predefined skill ratings as ground-truth training labels. To provide a relevant reference, we implemented a regression model to predict sonographer skill from ultrasound video frames. For this supervised approach, we utilised years of experience as the skill indicator, and each sonographer's years of experience were normalised to a score in the range of $[0,1]$ to match the scale of the output score from our proposed skill predictor. The framework adopted ResNet18 a benchmark quality ResNet18 as the regression model, which was trained using $\mathcal{D}^{train}$ and evaluated on $\mathcal{D}^{test}$, the same data used for developing our proposed framework. The regression model was trained to predict skill scores from the same input ultrasound frame sequences. The network was trained for 800 epochs using the MSE loss function and the evaluation model was selected based on the best validation set performance.

\section{Results}\label{sec:results}

\subsection{Task predictor}
As presented in Table~\ref{res-seg-main}, the segmentation task predictor was evaluated using the Dice score and average symmetric surface distance (ASSD). The model with rank normalisation and $L^{\theta}_{min}$ achieved the highest Dice score and the lowest ASSD for all anatomical landmarks. The Dice score on HC and CSP were significantly higher than all other combinations of normalisation and $L^{\theta}$ (with all $p\leq0.018$). The Dice score on LV showed no significant improvement ($p=0.0722$) to the model with rank normalisation and $L^{\theta}_{top-20\%}$, but was significantly higher compared to all other models (with all $p<0.0001$). The ASSD on HC and LV were significantly lower than all other models (with all $p<0.0001$). For ASSD on CSP, the model showed statistically significant improvement to other models (with all $p<0.0001$) except for the model with the min-max normalisation and $L^{\theta}_{avg}$ ($p=0.412$).

Comparing rank and min-max as normalisation methods, rank outperformed min-max on both Dice and ASSD except for the model with $L^{\theta}_{min}$. For models with $L^{\theta}_{avg}$ and  $L^{\theta}_{top-20\%}$, the Dice scores using rank were higher than those using min-max (with $p\leq0.001$ on all anatomical structures), while ASSD of models using rank also outperforms those using min-max ($p<0.0001$ for all compared models, except for the model using $L^{\theta}_{avg}$ on CSP, where $p=0.412$). 
The observed performance gain from the rank-based normalisation may be explained by its expected robustness to extreme values, as previously discussed in Sect.~\ref{sec:method-opt-task}.


Among the three formulations of $L^{\theta}$, the lowest dice score and highest ASSD were observed from models with $L^{\theta}_{min}$ on a majority of anatomical structures, indicating an inferior performance. 
When using rank normalisation, both $L^{\theta}_{avg}$ and $L^{\theta}_{top-20\%}$ outperforms $L^{\theta}_{min}$ on all evaluation metrics and anatomical landmarks ($p<0.001$ for all pairs). When using min-max normalisation, compared to $L^{\theta}_{min}$, the model with $L^{\theta}_{avg}$ had higher Dice scores and lower ASSD for all anatomical landmarks, and statistical significance was shown for both Dice and ASSD across all landmarks ($p<0.001$ for all pairs). 
Although the model with $L^{\theta}_{top-20\%}$ was outperformed by $L^{\theta}_{min}$, the performance drop was not found statistically significant for the Dice score for HC and LV, as well as all metrics for the CSP.
$L^{\theta}_{min}$ uses only one frame with the lowest loss to represent the whole sequence, whereas $L^{\theta}_{avg}$ and $L^{\theta}_{top-20\%}$ captures the average of the whole sequence or the skill predictor selected frames respectively, where the latter is a partial sequence representing the whole.

In terms of different anatomical landmarks, HC achieved the best performance for both Dice and ASSD, which could be because, 1) HC is relatively large in the frame as shown in Fig.~\ref{fig:label}, where research has shown correlation exists between the dice score and the size of the region-of-interest \cite{eelbode2020optimization}; and 2) there are more HC annotations in the training dataset as listed in Tab.~\ref{tab:dataset}. The Dice scores were similar for LV and CSP, however, CSP predictions generally had a smaller ASSD than LV.

\subsubsection{Effect of minibatch size}
Results for task predictor performance when training on different minibatch sizes are presented in Tab.~\ref{res-seg-bs}. For models using rank normalisation, the model performance dropped for all evaluation metrics when the minibatch size increased. The best performance was achieved for all anatomical landmarks with a minibatch size of $8$ for all evaluation metrics, and statistical significance was found between each of the first and second best results (with all $p<0.0001$). For models using min-max normalisation, comparable results were achieved with a minibatch size of $8$ and $32$. Similar to using rank normalisation, using a larger minibatch size of $64$ resulted in worse performance for all evaluation metrics.
This observed negative performance correlation to the minibatch size may be due to the change of the skill definition, as the normalisation method introduced in Sect.\ref{sec:method-opt-task} was designed based on minibatch. The practical interpretation of this difference remains an open question for specific future clinical studies.

\begin{table}[]
    \centering
    \caption{Skill predictor performance on different normalisation and formulations of $L^{\theta}$. Best values are marked in bold. ${top1}={sp}$ and ${top5}\supset{sp}$ indicates the ratio of test scans where skill predictor-rated top $1$ and top $5$ sequences contain the sonographer-selected standard plane respectively. $Dist.$ represents the average distance between the sequence assigned with the highest skill score and the sonographer-selected standard plane. }
    \resizebox{0.5\textwidth}{!}{%
    \begin{tabular}{llllllll}
    \toprule
        Norm & $L^{\theta}$ & MSE & $R_{top1}$ & $R_{top5}$ & ${top1}={sp}$ & ${top5}\supset{sp}$ & \begin{tabular}[c]{@{}c@{}}$Dist.$ \\ (s)\end{tabular}  \\ \midrule
        rank & $L^{\theta}_{avg}$ & \textbf{0.0240} & 0.76 & 0.88 & 0.16 & 0.56 & 1.97  \\ 
        rank & $L^{\theta}_{min}$ & 0.0329 & \textbf{0.96} & 0.92 & 0.20 & 0.60 & 2.64  \\ 
        rank & $L^{\theta}_{top-20\%}$ & 0.0360 & 0.92 & 0.92 & 0.16 & 0.60 & 2.00  \\ 
        min-max & $L^{\theta}_{avg}$ & 0.0251  & 0.88 & 0.92 & 0.24 & 0.68 & 1.86  \\ 
        min-max & $L^{\theta}_{min}$ & 0.0295  & 0.88 & \textbf{1.00} & 0.32 & 0.56 & 2.21  \\ 
        min-max & $L^{\theta}_{top-20\%}$ & 0.0297  & 0.88 & \textbf{1.00} & 0.16 & 0.56 & 3.19 \\ \bottomrule
    \end{tabular}
    }
    \label{res-skill-main}
\end{table}

\begin{table}[]
    \centering
    \caption{Effect of minibatch size on the skill predictor. Best values are marked in bold. ${top1}={sp}$ and ${top5}\supset{sp}$ indicates the ratio of test scans where skill predictor-rated top $1$ and top $5$ sequences contain the sonographer-selected standard plane respectively. $Dist.$ represents the average distance between the sequence assigned with the highest skill score and the sonographer-selected standard plane.}
    \resizebox{0.5\textwidth}{!}{%
    \begin{tabular}{llllllll}
    \toprule
        Norm & \begin{tabular}[c]{@{}c@{}}minibatch \\ size\end{tabular} & MSE & $R_{top1}$ & $R_{top5}$ & ${top1}={sp}$ & ${top5}\supset{sp}$ & $Dist.$(s) \\ \midrule
        rank & 8 & 0.0359 & 0.84 & 0.84 & 0.12 & 0.68 & 1.68  \\ 
        rank & 16 & 0.0329 & \textbf{0.96} & 0.92 & 0.20 & 0.60 & 2.64  \\ 
        rank & 32 & 0.0229 & 0.84 & \textbf{0.96} & 0.20 & 0.80 & 1.69  \\ 
        rank & 64 & \textbf{0.0118} & \textbf{0.96} & \textbf{0.96} & 0.16 & 0.52 & 2.68  \\ \midrule
        min-max & 8 & 0.0302 & 0.88 & 0.95 & 0.12 & 0.60 & 2.64  \\ 
        min-max & 16 & 0.0295 & 0.88 & \textbf{1.00} & 0.32 & 0.56 & 2.21  \\ 
        min-max & 32 & 0.0320 & 0.80 & 0.88 & 0.16 & 0.44 & 2.88  \\ 
        min-max & 64 & \textbf{0.0131} & \textbf{0.92} & 0.92 & 0.20 & 0.60 & 2.74 \\ \bottomrule
    \end{tabular}
    }
    \label{res-skill-bs}
\end{table}

\subsubsection{Effect of choice of $L^{\theta}_{top-m\%}$}
We conducted an experiment to compare the effect of different values of $m\%$ for $L^{\theta}_{top-m\%}$. The results for the segmentation task predictor are presented in Tab.~\ref{res-seg-top}. As shown in the table, for models using min-max normalisation, task predictor with $L^{\theta}_{top-30\%}$ achieved the highest Dice score and the lowest ASSD on all anatomical structures. For models using rank normalisation, task predictor with $L^{\theta}_{top-30\%}$ obtained the highest Dice scores on HC and CSP, while the lowest ASSDs were achieved on HC and LV.

\subsubsection{Effect of sequence length and sampling rate}
We compared the effect of using different combinations of frame sampling rate and sequence length $\tau$, which ranged between a total of $1/3$ to $2$ seconds. As shown in Tab.~\ref{res-seg-seqlen}, an input frame sequence sampled at 15 frames per second with a sequence length of 10 frames achieved the highest Dice score and lowest ASSD for all landmarks except for the CSP.

\subsection{Skill predictor}
\begin{table}[]
    \centering
    \caption{Skill predictor evaluation for different $L^{\theta}_{top-m\%}$. Best values are marked in bold. ${top1}={sp}$ and ${top5}\supset{sp}$ indicates the ratio of test scans where skill predictor-rated top $1$ and top $5$ sequences contain the sonographer-selected standard plane respectively. $Dist.$ represents the average distance between the sequence assigned with the highest skill score and the sonographer-selected standard plane.}
     \resizebox{0.5\textwidth}{!}{%
    \begin{tabular}{llllllll}
    \toprule
        Norm & $L^{\theta}_{top-m\%}$ & MSE & $R_{top1}$ & $R_{top5}$ & ${top1}={sp}$ & ${top5}\supset{sp}$ & $Dist$  \\ \midrule
        rank  & $L^{\theta}_{top-20\%}$ & 0.036 & \textbf{0.88} & \textbf{1.00} & 0.16 & 0.56 & 3.20  \\ 
        rank  & $L^{\theta}_{top-30\%}$ & 0.036 & 0.80 & 0.92 & 0.16 & 0.44 & 1.86  \\ 
        rank  & $L^{\theta}_{top-40\%}$ & \textbf{0.029} & 0.84 & 0.96 & 0.12 & 0.56 & 1.94  \\ 
        rank  & $L^{\theta}_{top-50\%}$ & 0.032 & 0.76 & 0.92 & 0.20 & 0.60 & 1.62 \\ \midrule
        min-max & $L^{\theta}_{top-20\%}$ & 0.0297  & \textbf{0.92} & \textbf{0.92} & 0.16 & 0.60 & 2.00  \\ 
        min-max & $L^{\theta}_{top-30\%}$ & 0.0342  & 0.76 & 0.88 & 0.12 & 0.44 & 2.67  \\ 
        min-max & $L^{\theta}_{top-40\%}$ & \textbf{0.0272}  & 0.80 & 0.88 & 0.12 & 0.60 & 2.02  \\ 
        min-max & $L^{\theta}_{top-50\%}$ & 0.0313  & 0.88 & \textbf{0.92} & 0.16 & 0.64 & 1.76 \\ \bottomrule
    \end{tabular}
    }
    \label{res-skill-avg}
\end{table}

To evaluate the performance of the skill predictor, we report the MSE between the skill predictor output and the performance of the task predictor, as well as the ratio of performance improvement, $R_{top1}$ and $R_{top5}$. 
For comparison to sonographer selection, we counted the number of cases where the skill predictor-rated top $1$ or top $5$ sequences that contain the standard plane selected by the sonographers and calculated the ratio of those cases to the total number of scans tested, noted as ${top1}={sp}$ and ${top5}\supset{sp}$, respectively. 
In addition, the average distance in seconds between the sequence assigned with the highest skill score and the sonographer-selected plane was also reported, noted as $Dist$. 

\begin{table}[]
\centering
\caption{Comparison of skill predictor performance on different sequence lengths and sampling rates. Best values are marked in bold. ${top1}={sp}$ and ${top5}\supset{sp}$ indicates the ratio of test scans where skill predictor-rated top $1$ and top $5$ sequences contain the sonographer-selected standard plane respectively. $Dist.$ represents the average distance between the sequence assigned with the highest skill score and the sonographer-selected standard plane.}
\label{res-skill-len}
\resizebox{0.5\textwidth}{!}{%
\begin{tabular}{llllllll}
\toprule
\begin{tabular}[c]{@{}c@{}}Frame rate \\ (fps)\end{tabular} & $\tau$ & MSE & $R_{top1}$ &         $R_{top5}$ & ${top1}={sp}$ & ${top5}\supset{sp}$ & $Dist$                       \\ \midrule 
        30 & 10 & 0.0295  & 0.88 & \textbf{1.00} & 0.32 & 0.56 & 2.21  \\ 
        30 & 20 & \textbf{0.0292} & 0.92 & 0.92 & 0.36 & 0.88 & 1.94  \\ 
        30 & 30 & 0.0389 & \textbf{0.96} & 0.92 & 0.52 & 0.96 & 1.36  \\ 
        15 & 10 & 0.0422  & 0.92 & 0.96 & 0.44 & 0.80 & 1.48  \\ 
        15 & 20 & 0.0338 & 0.92 & 0.96 & 0.52 & 1.00 & 1.45  \\
        10 & 10 & 0.0328  & 0.92 & 0.96 & 0.44 & 1.00 & 1.62  \\
        10 & 20 & 0.0330 & 0.92 & 0.64 & 0.92 & 1.00 & 0.09 \\
\bottomrule
\end{tabular}%
}
\end{table}

As shown in Tab~\ref{res-skill-main}, the model with rank normalisation and $L^{\theta}_{avg}$ obtained the lowest MSE, indicating the highest similarity between the predicted skill score and task predictor performance on the input sequence. 
For the ratio $R_{top1}$, $96\%$ of the top $1$ sequences attained increased Dice scores when using the model with rank normalisation and $L^{\theta}_{min}$, which was the highest ratio of performance gain. 
For $R_{top5}$, both $L^{\theta}_{min}$ and $L^{\theta}_{top-20\%}$ with min-max normalisation models provided dice score improvement on $100\%$ of the test data.
The ratio of task performance improvement is an indicator showing that leveraging the frame sequences selected by the skill predictor enhanced the predictive capabilities of the lower-level task predictor. 
The results in Tab~\ref{res-skill-main} indicate that, among the models assessed, $5$ out of $6$ task predictor models showcased performance improvement on at least $88\%$ of test scans. These improvements were observed when using the highest-rated (top 1) sequence selected by the skill predictor, which was deemed most conducive to the task predictor. 
All models demonstrated performance improvements on at least $88\%$ of the test scans when employing the top $5$ candidate sequences.
The results suggest that the proposed models were able to select the frames that are more conducive to the target clinical task predictor.

\subsubsection{Effect of minibatch size}
Results comparing model performance using different minibatch sizes are presented in Tab.~\ref{res-skill-bs}. Models that used a large minibatch size of $64$ obtained the lowest MSE regardless of using rank or min-max normalisation methods. The highest $R_{top1}$ was achieved using a minibatch size of $64$. A possible explanation for this finding is that a larger minibatch size contains more sequences leading to a better representation of the comparison or ranking of skill.

\subsubsection{Effect of choice of $L^{\theta}_{top-m\%}$}
The effect of various values of $m\%$ of the selection function $L^{\theta}_{top-m\%}$ was investigated, with results presented in Tab.~\ref{res-skill-avg}. 
The lowest MSE was observed with a selection function of $L^{\theta}_{top-40\%}$ for models with both normalisation methods.
However, the highest ratio of $R_{top1}$ and $R_{top5}$ was achieved with $L^{\theta}_{top-20\%}$.

\subsubsection{Effect of sequence length and sampling rate}
We compared the impact of various combinations of frame sampling rates and sequence lengths ($\tau$), as presented in Tab.~\ref{res-skill-len}. 
The skill predictor model, utilising input sequences of 20 frames sampled at 30 frames per second, achieved the lowest MSE. 
In terms of performance improvement rates, the model using 30-frame and 10-frame inputs exhibited the highest ratios for $R_{top1}$ and $R_{top5}$ respectively, both sampled at 30 frames per second.
\begin{figure}[htb]
     \centering
     \includegraphics[width=0.5\textwidth]{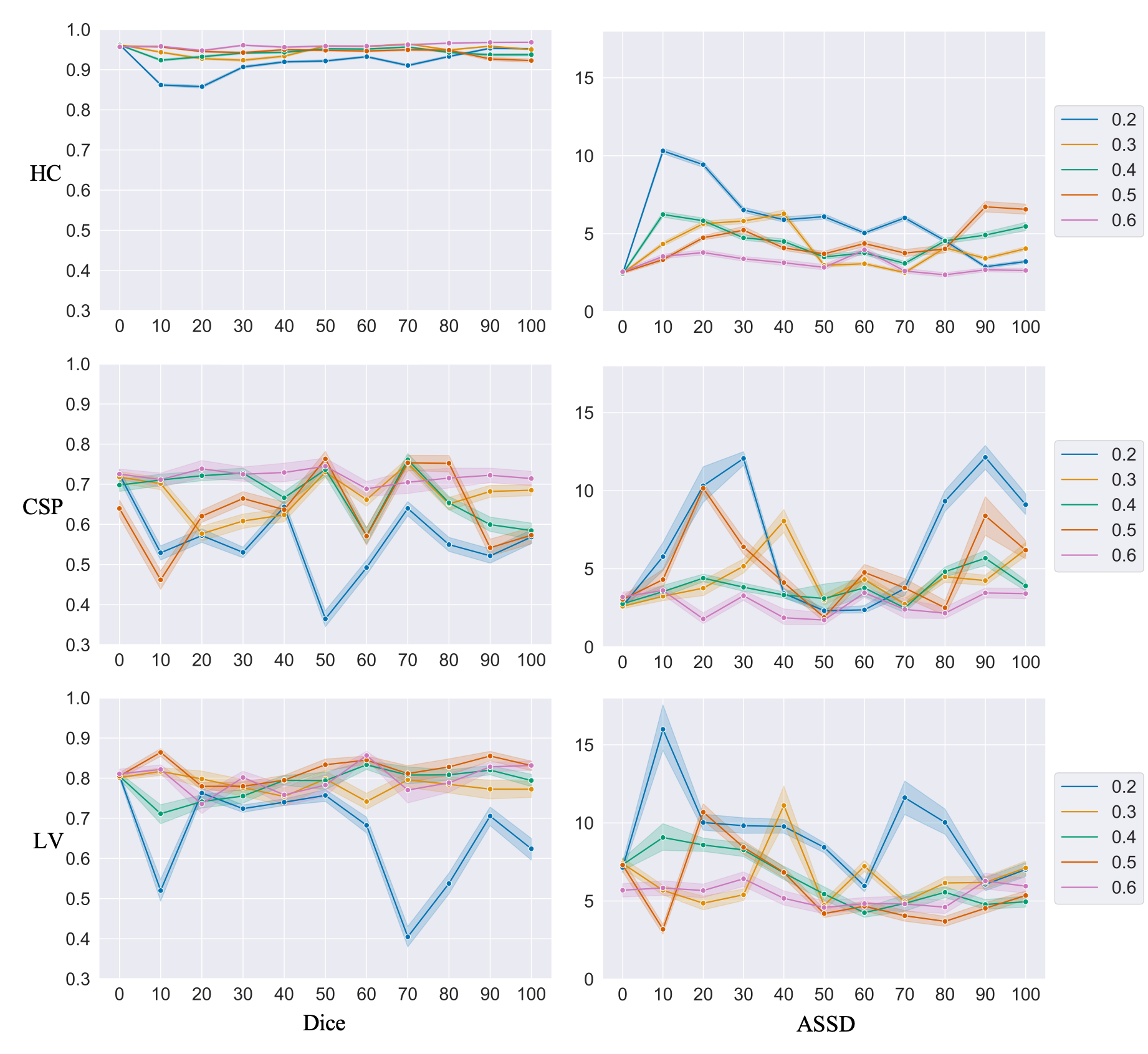}
     \caption{Meta evaluation results of the task predictor. Solid lines with different colours represent model performance fine-tuned using different ratios of the test dataset, while the shades indicate the variance.}
     \label{fig:meta-seg}
 \end{figure}

 \subsection{Meta evaluation}
We performed meta evaluation to understand the model performance when fine-tuned on different sizes of the meta test dataset. Models were fine-tuned for $\{10, 20, ..., 100\}$ epochs using $\{20\%, 30\%, ..., 60\%\}$ of the test dataset.

\subsubsection{Task predictor}
We plotted how the task predictor performance changed when finetuned on different percentages of the test dataset with different epochs, including the Dice score and ASSD on HC, CSP, and LV, as shown in Fig~\ref{fig:meta-seg}.
When using $20\%$ of the test dataset, the ASSD of CSP and LV obtained lower values during the fine-tuning process, compared to testing on the same data using the best model selected during training. When fine-tuning on $30\% - 60\%$ of the test dataset, in most cases models achieved performance gain for both Dice and ASSD, except for HC when using $40\%$ and $50\%$ of the test data. In addition, best performance were achieved mostly when fine-tuning for $50-70$ epochs.
The results indicate that the finetuning is beneficial for the task predictors, by producing segmentation predictions with higher Dice scores and lower ASSDs with less than 100 epochs. Moreover, 30\% of the test data is able to achieve comparable performance compared to finetuning on higher ratios of the test data.

\begin{figure*}[!h]
    \centering
    \includegraphics[width=0.9\textwidth]{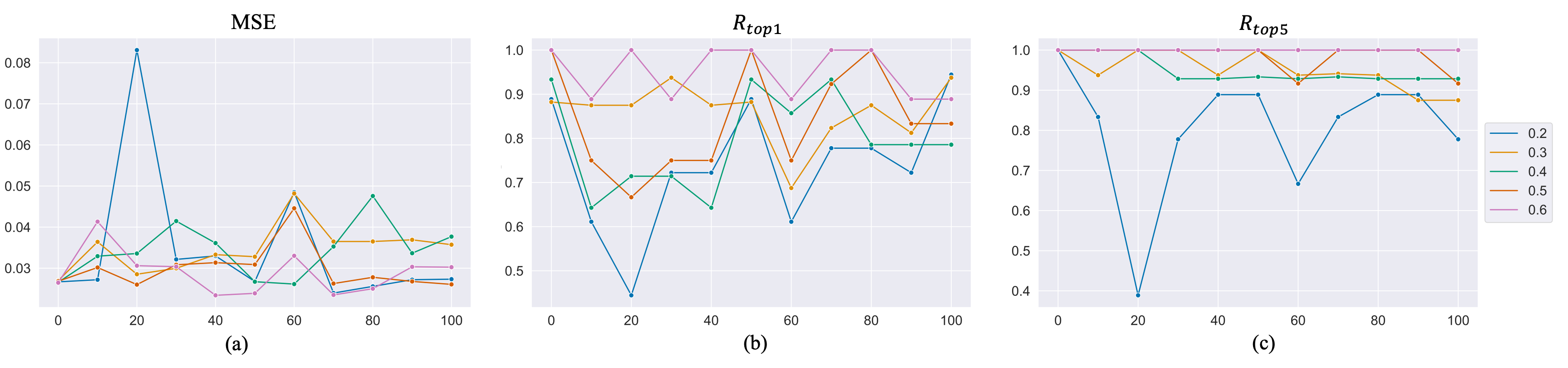}
    \caption{Meta evaluation results of the skill predictor. Different colours represent models fine-tuned using different ratios of the test dataset.}
   \label{fig:meta-skill}
\end{figure*}

\begin{figure*}[!h]
    \centering
    \includegraphics[width=0.9\textwidth]{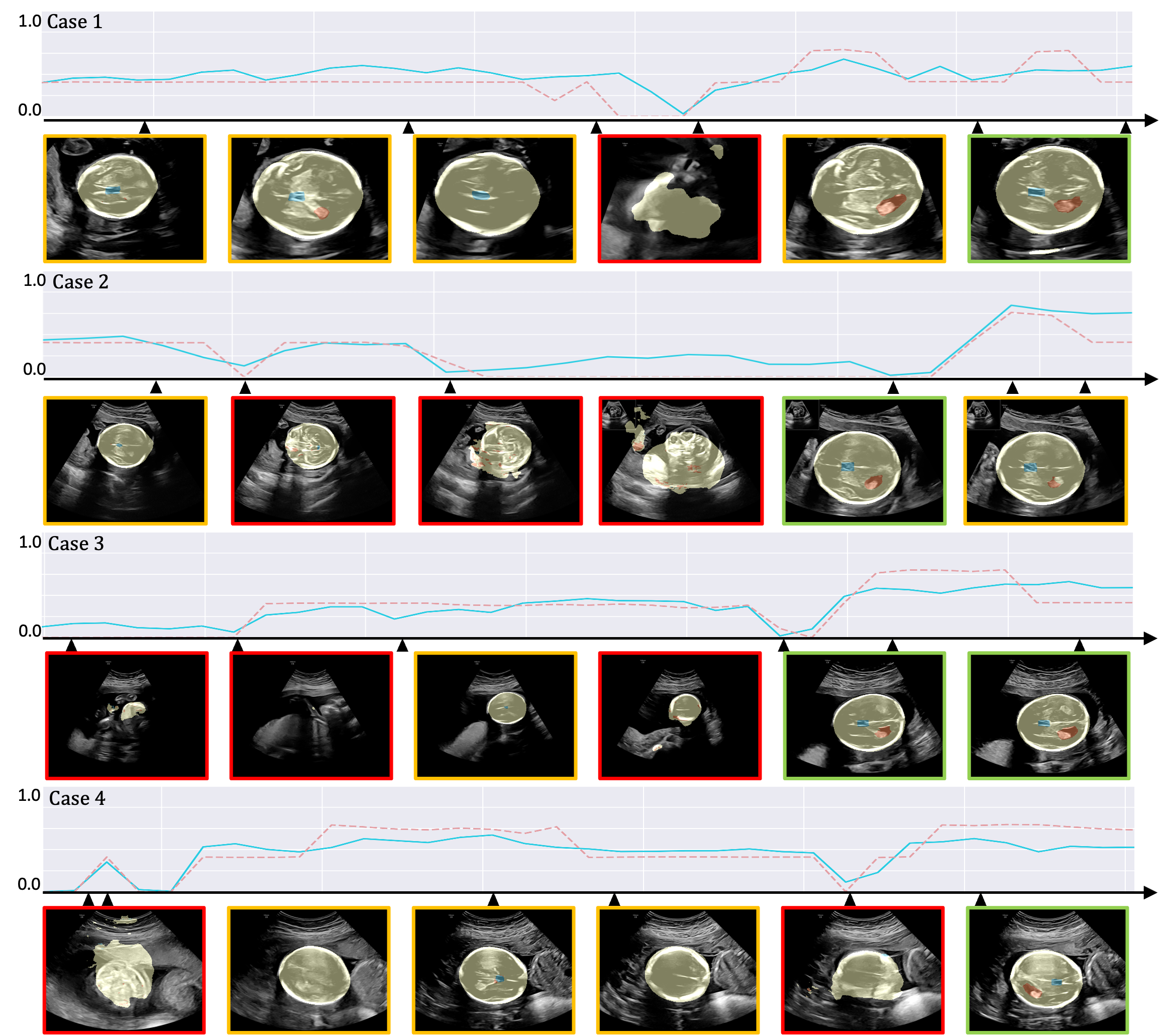}
    \caption{Four example scans from four sonographers with different years of experience plotted along the time, with the time synchronised skill assessment scores. The blue solid line and the red dashed line represent the skill score given by the skill predictor and the evaluation metric for the task prediction. Corresponding ultrasound frames at the time points marked by a black triangle are presented below the line plot chronologically.}
   \label{fig:res-score}
\end{figure*}
 
\subsubsection{Skill predictor}
Figure~\ref{fig:meta-skill} presents how MSE and ratio of performance improvement $R_{top1}$ and $R_{top5}$ change along the fine-tuning epochs using different ratios of the test dataset. 
After finetuning, models except for using $30\%$ of the test dataset were able to achieve a lower MSE score, typically after finetuning for 70 epochs.
When assessing the improved performance rate of top 1 selected sequences, a higher ratio was achieved when finetuning on $20\%$ and $30\%$ of the test data, while for others the same ratio was obtained as the models without finetuning. Note that for models finetuned on $50\%$ and $60\%$ of the test set, a $100\%$ of performance improvement was achieved on the model without finetuning therefore no higher ratio can be obtained. 
For the ratio of improved performance on the top 5 selected sequences, all models achieved $100\%$ of performance improvement before finetuning, and 4 out of 5 models were able to achieve the same after finetuning.

\subsection{Qualitative case studies}
In Figure~\ref{fig:res-score}, we present four example test scan visualisations, performed by four sonographers with 2, 5, 6, and 7 years of experience respectively. For each Case the blue solid line indicates the skill score given by the skill predictor, while the red dashed line represents the evaluation metric for the task predictor. Several interesting points were selected manually and plotted along the time axis, marked as black triangles, and the ultrasound frames at those interesting time points are presented below the line plot chronologically. 

The timestamps when the standard transventricular plane appears are indicated by the green boxes. At this timestamp, the segmentation task predictor accurately predicts all three anatomical landmarks, while the skill predictor gives a high score indicating the acquired frames are associated with good skill.

When a sonographer is approaching a standard plane, the ultrasound frames may present a clear skull but only one or two landmarks are visible. In such instances, the task predictor is able to give a prediction that only includes the shown structures. In Fig.~\ref{fig:res-score} the yellow boxes highlight these frames where only the HC, or HC and CSP, are segmented. In these frames, the skill predictor produces a score lower than that observed in the standard plane scenario.

We also observed that in some poorly acquired frames, shown with red boxes in the figure, the fetal head was not clearly visible. In these cases, the task predictor is unable to identify any anatomical landmarks, resulting in nearly empty prediction masks or poorly shaped masks. These frames are rated with low scores by the skill predictor.

Years of experience has been used as a skill indicator in conventional skill assessment methods and automatic methods \cite{zago2020educational,wang2020differentiating}. The four cases in Fig.~\ref{fig:res-score} were scans performed by sonographers with increasing years of experience, yet all demonstrated fluctuating scores along the scanning process. This suggests that using years of experience is not sufficient to evaluate the skills demonstrated in individual scans at specific time points. Furthermore, this limitation is more evident when training a deep learning model using years of experience as the supervision signal. As demonstrated in the qualitative comparison in the Appendix, the supervised baseline was unable to generate skill scores that correspond to the clinical relevance of the acquired ultrasound frames, in this case, finding the appropriate plane for the head circumference measurement task.
Instead, our proposed model evaluates the acquired image sequences in relation to their ability to support the clinical task predictor. In a clinical setting, these scores could guide sonographers by indicating how well the currently acquired image supports the clinical objective, helping them determine whether adjustments to the scanning plane are needed. This shows that the proposed method offers new insights into sonographer skill assessment by shifting the focus from general experience to task-specific performance.

\section{Discussion}
In this work, we have introduced a framework that assesses ultrasound skills using a deep learning model-automated clinical task. Once trained, the proposed skill predictor can be used independently of the task predictor, providing real-time scores to guide a sonographer on whether the acquired ultrasound images are sufficient for the selected clinical tasks. In addition, the task predictor can serve as a clinical tool, for data-acquisition-dependent task, in addition to providing interpretation of the decisions made by the skill predictor. Training a skill predictor to predict skill scores rather than calculating the performance of the task predictor is essential, such that the ground-truth labels for the clinical task are not required, which are often unavailable when acquiring data from new ultrasound scans or in new clinical tasks.

During our extensive experiments, we did not encounter scenarios where the task predictor performed poorly that consequently causing the failure of the skill predictor. However, as the two networks train from scratch simultaneously, adding warm up epochs to train the task predictor could potentially help avoid training collapse or improve training efficiency. 

One limitation of this work is that the skill predictor is designed to be task-specific, which indeed means that the framework requires ground-truth labels from specific tasks of interest and retraining for different clinical tasks. However, the proposed method allows the model to be tailored to the unique requirements of each task, ensuring optimal skill prediction and relevance in different clinical applications.
Another limitation of this work is that pixel size (in mm) was not available for our data which restricted us to report ASSD in pixel units rather than physical units. Although this may affect the interpretability of ASSD, it does not impact the training or evaluation of the skill predictor. Future work could consider the impact of pixel spacing to improve clinical relevance. Despite this limitation, conclusions regarding the robustness and utility of the proposed skill predictor remain unaffected.
To further validate the clinical value of the proposed framework, prospective clinical studies can be designed to compare the outcome on selected clinical tasks using frames acquired with versus without the assistance of our skill assessment model.
In addition, comparing other methods from bi-level optimisation or meta-learning might form an interesting part of future work, but are not hypothesised to significantly change the key findings in this work.

\section{Conclusion}

In conclusion, this work proposed a bi-level optimisation framework to assess fetal ultrasound skills by a specific clinical task, without using any predefined skill ratings. The framework consists of an anatomical landmark segmentation network and a skill assessment network, which are optimised jointly by refining the two networks simultaneously. We validated the proposed method on a real-world fetal ultrasound dataset, with three segmentation tasks. The quantitative experiment results and the discussed case studies collectively demonstrate the feasibility of selecting ultrasound frames associated with a skill score that better represents the optimised task performance, a characteristic in skill assessment that is considered ideal in this study.

\appendices


\section*{Acknowledgment}
This work was supported by the European Research Council (ERC), Project PULSE, under Grant ERC-ADG-2015694581, the National Institute for Health and Care Research (NIHR) Oxford Biomedical Research Centre, and an EPSRC Turing AI World Leading Researcher Fellowship (EP/X040186/1). This work was also supported by the Wellcome/EPSRC Centre for Interventional and Surgical Sciences [203145Z/16/Z].

\section*{References}

\bibliographystyle{IEEEtran}  
\bibliography{ref.bib}

\section*{Appendix}
\begin{figure*}
\label{app:fig}
\centering
\includegraphics[width=1\textwidth]{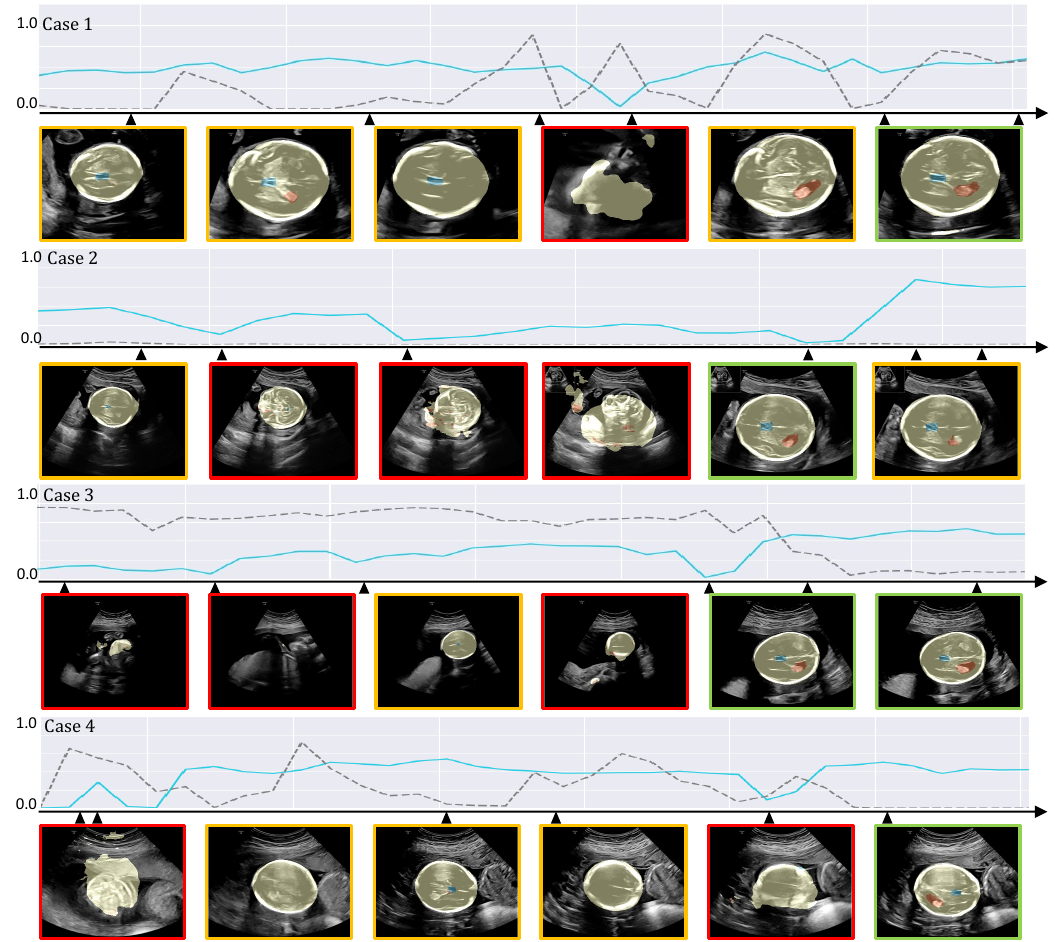}
\caption{Four example scans from four sonographers with different years of experience plotted along the time, with the time synchronised skill assessment scores. The blue solid line and the gray dashed line represent the skill score given by the proposed skill predictor and the comparison supervised-learning regression model. Corresponding ultrasound frames at the time points marked by a black triangle are presented below the line plot chronologically.} 
\end{figure*}

\end{document}